\documentclass[10pt,twocolumn,letterpaper]{article}

\usepackage{iccv}
\usepackage{times}
\usepackage{epsfig}
\usepackage{graphicx}
\usepackage{amsmath}
\usepackage{amssymb}
\usepackage{xcolor}         
\usepackage{multirow}
\usepackage[export]{adjustbox}
\usepackage{booktabs}
\usepackage{float}
\usepackage{subfig}
\usepackage{enumitem}
\usepackage{comment}
\usepackage{color, colortbl}

\usepackage[pagebackref=true,breaklinks=true,letterpaper=true,colorlinks,bookmarks=false,citecolor=citecolor]{hyperref}
\definecolor{citecolor}{HTML}{0071BC}
\definecolor{dgreen}{RGB}{1,150,74}

\newcommand\cyan[1]{\textcolor{cyan}{#1}}
\newcommand\red[1]{\textcolor{red}{#1}}
\newcommand\dg[1]{\textcolor{dgreen}{#1}}

\newcommand{\tablestyle}[2]{\setlength{\tabcolsep}{#1}\renewcommand{\arraystretch}{#2}\centering\footnotesize}
\def\name{CTRL }
\def\namenospace{CTRL}

\def\slogannospace{Once Detected, Never Lost}
\def\sloganlower{once detected, never lost}

\iccvfinalcopy 

\def\iccvPaperID{2093} 

\begin{document}

\title{\slogannospace:\\Surpassing Human Performance in Offline LiDAR based 3D Object Detection}

\author{Lue Fan\\
CASIA\\
\and
Yuxue Yang\\
CASIA\\
\and
Yiming Mao\\
HNU; TuSimple\\
\and
Feng Wang\\
TuSimple\\
\and
Yuntao Chen\\
CAIR, HKISI, CAS\\
\and
Naiyan Wang\\
TuSimple\\
\and
Zhaoxiang Zhang\\
CASIA\\
\and
{
\tt\small
\{fanlue2019, yangyuxue2023, zhaoxiang.zhang\}@ia.ac.cn \;mym0729@hnu.edu.cn} \\
{\tt\small
\{feng.wff, chenyuntao08, winsty\}@gmail.com
}
}

\maketitle
\ificcvfinal\thispagestyle{empty}\fi

\begin{abstract}
This paper aims for high-performance offline LiDAR-based 3D object detection.
We first observe that experienced human annotators annotate objects from a track-centric perspective.
They first label the objects with clear shapes in a track, and then leverage the temporal coherence to infer the annotations of obscure objects.
Drawing inspiration from this, we propose a high-performance offline detector in a track-centric perspective instead of the conventional object-centric perspective.
Our method features a bidirectional tracking module and a track-centric learning module.
Such a design allows our detector to infer and refine a complete track once the object is detected at a certain moment.
We refer to this characteristic as ``on\textbf{C}e detec\textbf{T}ed, neve\textbf{R} \textbf{L}ost'' and name the proposed system \textbf{\namenospace}.
Extensive experiments demonstrate the remarkable performance of our method, surpassing the human-level annotating accuracy and the previous state-of-the-art methods in the highly competitive Waymo Open Dataset without model ensemble.
The code will be made publicly available at \url{https://github.com/tusen-ai/SST}.
\end{abstract}

\section{Introduction}
\begin{figure}[t]
    \centering
    \includegraphics[width=0.95\linewidth]{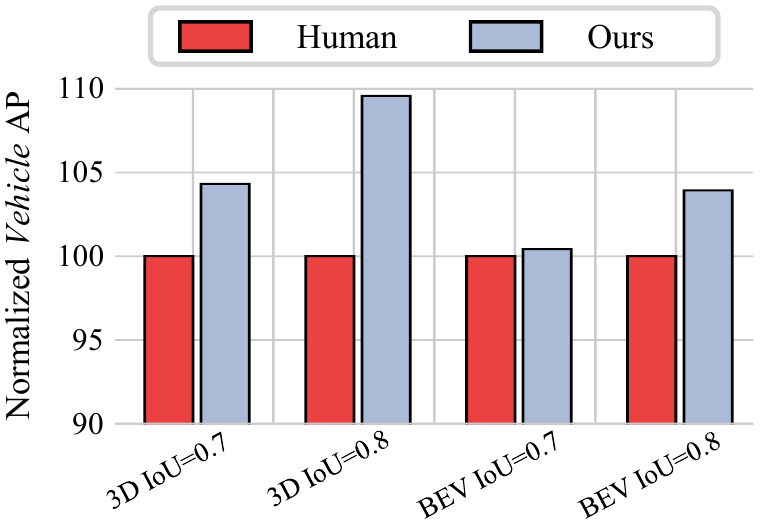}
    \caption{
    Our method consistently outperforms human performance in Waymo Open Dataset.
    Human performance is reported by 3DAL~\cite{offboard}.
    We normalize the AP of human performance to 100 for better visualization.
    }
     \vspace{-3mm}
    \label{fig:fig1}
\end{figure}
As a fundamental task in autonomous driving, 3D object detection has made great progress in recent years, in both LiDAR-based detectors~\cite{second, pointpillar, centerpoint, pvrcnn} and image-based detectors~\cite{detr3d, bevdet, bevformer, petr}.
Such success is primarily attributed to the data-driven paradigm, which requires a massive amount of labeled data.
As a result, there is growing interest in developing a high-performance offline detector for auto-labeling.

To achieve a high-performance offline detector, we first analyze the behavior of human annotators in a standard sequence labeling process. 
We find that humans annotate objects from a track-centric perspective, meaning that they utilize the temporal motion cues of objects to achieve precise labeling.
In particular, annotators first label easy samples in a sequence, and then use temporal cues to propagate these high-confidence labels to other time steps, when the objects may contain very few points and are hard to be accurately labeled individually.
In this process, once an object is detected in a certain moment, human annotators continuously track that object and predict its movements over time, even if there are some periods when the object is temporarily invisible.
An important fact is exploited here: \emph{a detected object will not disappear unless it moves outside the perception range.} 
Inspired by human labeling behavior, we develop a track-centric offline detector that aims to achieve two objectives:
1) Accurate labeling of well-observed tracks.
2) For tracks containing only a few high-quality frames, our detector propagates the predictions in high-quality frames to low-quality frames.
We refer to this key property of the detector as ``\emph{onCe detecTed, neveR Lost}'' and accordingly name our method \textbf{\namenospace}.

In order to implement our vision, we first utilize a base detector to obtain detection results.
Then we propose a bidirectional tracking module.
During the tracking process, a simple motion model is adopted to fill in missing frames and bidirectionally extend the track, greatly extending the life cycle of the track.
The bidirectional tracking offers the detector an opportunity to catch obscure objects which might be totally ignored by the base detector.
However, such a heuristic frame filling or extension cannot obtain precise poses of objects.
Therefore, we develop a track-centric learning module for refinement.
In the design of this module, we follow a track-centric principle: tracks, rather than objects, should be regarded as first-class citizens in the workflow.
This module takes all points and all proposals in the entire track as input, and refines all their poses simultaneously.
In addition, after the refinement, we could optionally optimize the track coherence via a temporal coherence optimization module if needed.
Particularly, we manually specify a bounding box containing a clear object shape in a track, and then align other object shapes with the specified one by point cloud registration\cite{p2picp, multiway}.
We summarize our contributions as follows.
\begin{itemize}[leftmargin=*]
    \item Based on the behavior of human annotators, we propose an offline detection system \namenospace, following the philosophy of ``track-centric'' and ``\sloganlower''.
    \name boosts the performance of auto-labeling.
    \item Single-model \name outperforms the previous state-of-the-art offline detector and all the online detectors.
    It is worth emphasizing that among millions of vehicles, only 0.48\% of them would be completely missed by \namenospace.
    \item We carefully relabel some diverged cases between our predictions and official ground truths. Our results demonstrate that our method even surpasses the ground-truth accuracy provided by Waymo human annotators in those cases.
    \item We keep our methodology simple and clean, greatly simplifying the workflow and reducing the resource requirements of existing offline frameworks.
\end{itemize}

\section{Related Work}
\paragraph{LiDAR-based 3D Object Detection}
LiDAR-based 3D object detectors usually adopt three representation modes, namely point-based~\cite{pointnet,pointnet++,votenet,pointrcnn,3dssd}, voxel-based~\cite{second,pointpillar,pvrcnn,sst,fsd},
and range-image-based~\cite{lasernet, rangedet, rcd,rsn}. 
Recently, combining multi-frame point clouds becomes a prevalent approach since it can provide a more comprehensive representation than a single-frame point cloud.
Multi-frame detectors~\cite{centerformer, centerpoint, afdetv2, pvrcnnpp, 3dman} have demonstrated that the concatenation of multi-frame point clouds can significantly outperform the single-frame setting. 
In addition to the straightforward point concatenation, some methods develop more well-designed temporal fusion strategies.
MPPNet~\cite{mppnet} incorporates multi-frame feature encoding and interaction modules, which results in higher performance.
INT~\cite{int} builds a temporal feature bank for temporal information aggregation.
FSD++~\cite{fsd++} leverages temporal information to generate residual points, reducing the temporal redundancy of multi-frame detection.

 \vspace{-3mm}
\paragraph{LiDAR-based 3D Multi-object Tracking}
Due to the accurate distance measurements that LiDAR provides, most state-of-the-art 3D Multiple Object Tracking (MOT) algorithms adopt a ``tracking-by-detection'' paradigm~\cite{3dmotbaseline}. The MOT in 3D space is usually easier than that in the 2D image space, since the motion of objects in 3D is much easier to predict which is beneficial for data association.
These algorithms proposed various methods to enhance data association~\cite{simpletrack,gnn3dmot,centerpoint}, motion propagation~\cite{probtrack}, and object life cycle management~\cite{simpletrack,immortaltrack}. Specifically, 
Immortal Tracker~\cite{immortaltrack} demonstrates that even a simple motion model with (almost) forever track preservation could significantly reduce premature track termination. Their findings coincide with our observations of human annotators.
Recently, some of the latest methods incorporate learning-based algorithms to improve association with features from point clouds~\cite{spot,simtrack}. 
For example, SimTrack~\cite{simtrack} demonstrates an end-to-end trainable model for joint detection and tracking from raw point clouds by feature alignment. 

 \vspace{-3mm}
\paragraph{3D Object Automated Labeling}
In recent years, the annotation costs for training data-hungry models have increased significantly. 
Accurate auto-labeling can significantly reduce annotation time and cost~\cite{annotating}. 
For 3D object detection, 3DAL~\cite{offboard} used point cloud sequence data, which achieves significant gains compared to state-of-the-art onboard detectors and offboard baselines, and even matches human performance. 
Auto4D~\cite{auto4d} proposed an automatic annotation pipeline for generating accurate object trajectories in 3D space from LiDAR point clouds, achieving a 25\% reduction in human annotation workload. 
Numerous studies have proposed different settings to assist human annotators and mitigate the expenses associated with annotation~\cite{autolabeling,weakly3d,active3d,leverag3d}.


\begin{figure*}[t]
    \centering
    \includegraphics[width=\linewidth]{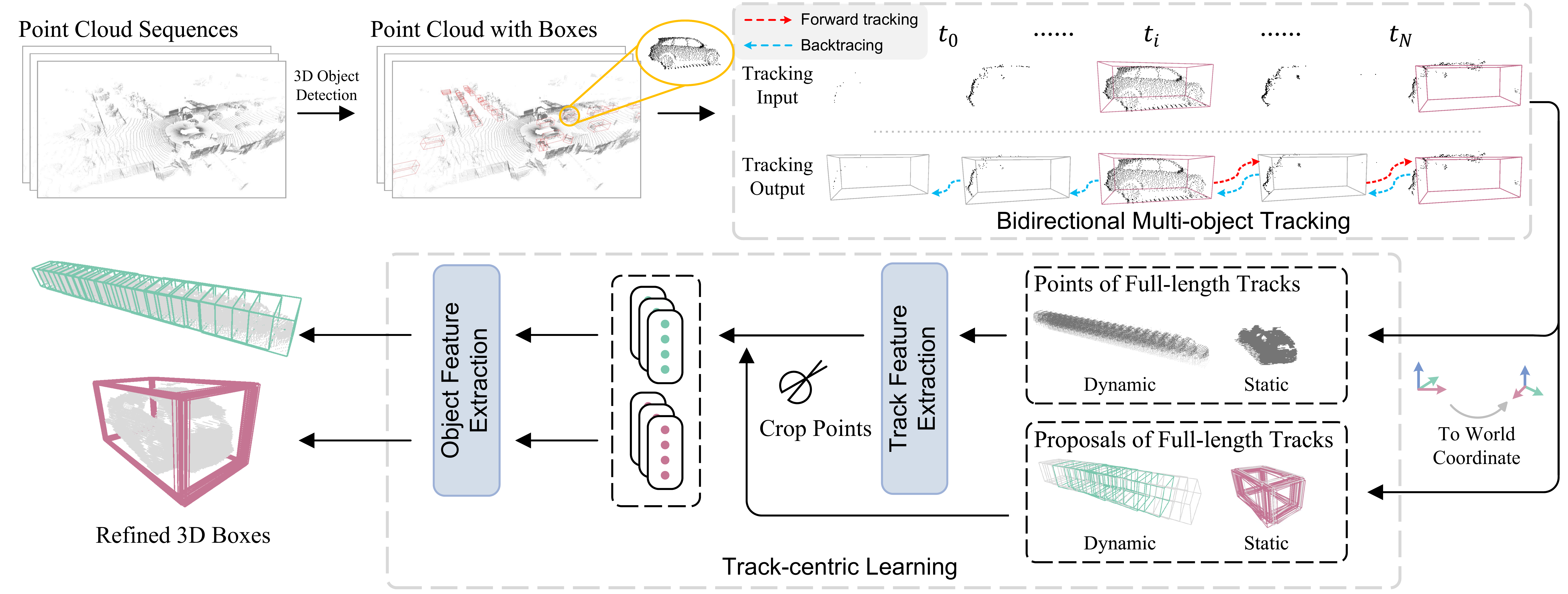}
    \caption{
    \textbf{Overall architecture of \namenospace.}
    (1) A base detector is utilized to generate basic detection results.
    (2) In the bidirectional tracking module, a forward tracking process is applied first to fill missing boxes and extend the track to the future, indicated by the \red{red} arrow. Then we backtrace to the start frame and extend the track to the past, indicated by the \cyan{blue} arrow.
    (3) The bidirectional extended tracks are sent into the track-centric learning module for refinement.
    }
    \label{fig:pipeline}
\end{figure*}
\section{\slogannospace}
\subsection{Base Detector}
We opt for the emerging FSD~\cite{fsd} as the base detector because of its proven effectiveness and ease of use.
Some modifications are made to make FSD better suit our needs.
(1) As our system is designed for offline use, in addition to the traditional multi-frame point cloud concatenation, we also incorporate future frames, which leads to a considerable performance improvement.
(2) In order to utilize longer temporal information without too much computational overhead, we employed a frame-skipping strategy. 
In practice, we sample 9 frames of data in total, i.e. $[t-8, t-6, \cdots, t, \cdots, t+6, t+8]$.
(3) To prevent overfitting, we employ the frame dropout strategy, in which half of the frames in the selected frames are randomly dropped with a probability of 20\%.

\subsection{Bidirectional Multi-object Tracking}
\paragraph{Forward Tracking}
Since the motion of objects in 3D space is much easier to predict than that in the 2D images, one simple idea to achieve ``never lost'' is to extend the track based on the motion model.
Recently, Immortal Tracker~\cite{immortaltrack} is designed to fit this principle, which predicts a pseudo-box by the motion model when a tracklet matches no observation in a time step.
Thus, the tracklets will never die unless the objects are out-of-range or the sequence ends.
With its ``immortal'' feature, it is able to associate broken tracks due to occlusion or missing detection in long term.
\paragraph{Backtracing}
The Immortal Tracker operates in a forward-only manner, which enables it to fill in missing boxes once an object has been initially detected.
However, the objects cannot appear from nothing. They should also exist before being detected.
Thus, we backtrace the tracks to the beginning of the tracks, and use the backward motion model to extend tracks into the past.
We illustrate this process in Figure~\ref{fig:pipeline}.

Here, it is necessary to point out that the positions and confidences of these added boxes are not accurate, leading to many false positives.
To address this issue, we propose a solution in the following \S\ref{sec:track_learning}.

\section{Track-centric Learning}
\label{sec:track_learning}
In this section, we propose a track-centric learning module to improve prediction quality.
This module takes full-length tracks as input which will be elaborated in \S\ref{sec:input}.
Its network structure consists of two major parts.
The first part utilizes the whole track for track-level feature extraction.
The second part extracts box-level features for proposal refinement.
We present these two parts in \S\ref{sec:feature_extraction}.
In \S\ref{sec:assign}, we propose a novel track-centric label assignment.
We also propose a couple of post-processing techniques to further improve the prediction quality in \S\ref{sec:post}.

\subsection{Organizing Track Input}
\label{sec:input}
\paragraph{Multiple-In-Multiple-Out}
In the previous object-centric designs~\cite{offboard, mppnet}, the model takes \emph{multiple} point clouds from nearby time steps to refine a \emph{single} proposal in every forward pass. 
For brevity, we name this as \emph{Multiple-In-Single-Out} (MISO).
Instead, in our track-centric design, the network takes a full-length track as input and simultaneously refines all proposals of the track in a single forward pass.
Correspondingly, we name this way as \emph{Multiple-In-Multiple-Out} (MIMO).
Specifically, in MIMO, we first expand proposals in a track by 2 meters in all three dimensions.
Then we crop the point clouds in every time step by the expanded proposals. 
The cropped points are transformed into the pose of the first frame of the track and concatenated together as network input.
MIMO enables two unique features of our methods.
\begin{itemize}[leftmargin=*]
    \item \textbf{Effective Training}
    With MIMO, our approach simultaneously applies supervision on every object in different time step in a single forward pass.
    Furthermore, the track features can be shared by all the objects in the track.
    Such dense supervision and feature sharing make the training very effective.
    \item \textbf{Computational Efficiency}
    In MISO, point clouds from multiple time steps are paired with a single proposal for input.
    Thus, the computational overhead grows along with the length of input point clouds.
    However, in MIMO design, all the point clouds in a track are concatenated and shared by all proposals.
    This characteristic allows for high efficiency and almost unlimited input length.
\end{itemize}
\vspace{-5mm}
\paragraph{Motion-state Agnostic}
The previous work~\cite{offboard} partitions all tracks by their motion states, and designs two different pipelines for the dynamic and static tracks, respectively.
However, we find it unnecessary and even harmful.
On one hand, such a partition reduces the diversity of training data, thus hindering generalization.
On the other hand, some categories may have unstable motion states, such as pedestrians.
Therefore, we don't distinguish the dynamic and static objects and treat them in a unified way throughout the whole pipeline, greatly simplifying the workflow.
\vspace{-3mm}
\paragraph{Timestamp Encoding}
To distinguish the points from different time steps, we attach timestamp encoding to the input points.
Although there are many options, we opt for the simplest one that uses their absolute frame ID in sequence as the time encoding.
Specifically, points from $i$-th frame will be attached with $0.01 \times i$ as its timestamp encoding.

\subsection{Sparse Feature Extraction}
\label{sec:feature_extraction}
Here we introduce the network structure for feature extraction and prediction. The network consists of a Sparse UNet\cite{parta2} backbone to extract features of the track points and a PointNet-like head to predict object parameters.
\vspace{-3mm}
\paragraph{Track Features}
Since we use full-length tracks as input, the spatial span of these tracks might be considerably large.
Thus, in practice, we empirically adopt an input space in the size of $[-256m, 256m] \times [-256m, 256m]$, which could cover the spatial spans of most tracks in WOD~\cite{wod}.
To handle such a large space, we turn to the emerging fully sparse architecture.
In particular, we first voxelize the point clouds of tracks. Then a Sparse UNet with aggressive downsampling is leveraged to extract the voxel features, which has sufficient receptive fields. Eventually, the voxel features are mapped back to their containing points by interpolation.
The obtained point features are utilized for object feature extraction in the following. 


\vspace{-2mm}
\paragraph{Object Features}
After the SparseUNet, we use proposals to crop the point features, where the proposals are expanded by a certain margin (e.g., 0.5 meters) to ensure the integrity of the object. Then an efficient PointNet is utilized to extract the proposal features, which contains several MLPs and max-pooling layers.
Note that a proposal in the time step $t$ will also crop the points from other time steps to get more additional information.
We append a special flag on the points in the current frame, making the model easier to recognize the current object shape, avoiding being confused by points from other time steps.

\subsection{Label Assignment}
\label{sec:assign}
Previous object-centric detectors~\cite{offboard, pvrcnn, parta2, fsd} match objects and ground-truth boxes with a certain metric (e.g., IoU) to assign labels.
Due to a lack of temporal constraints, proposals might be mistakenly assigned as negative.
In our method, global track information could be leveraged as strong priors for more robust label assignment. Therefore, we develop a track-centric label assignment.
\par
\begin{figure}
    \centering
    \includegraphics[width=\linewidth]{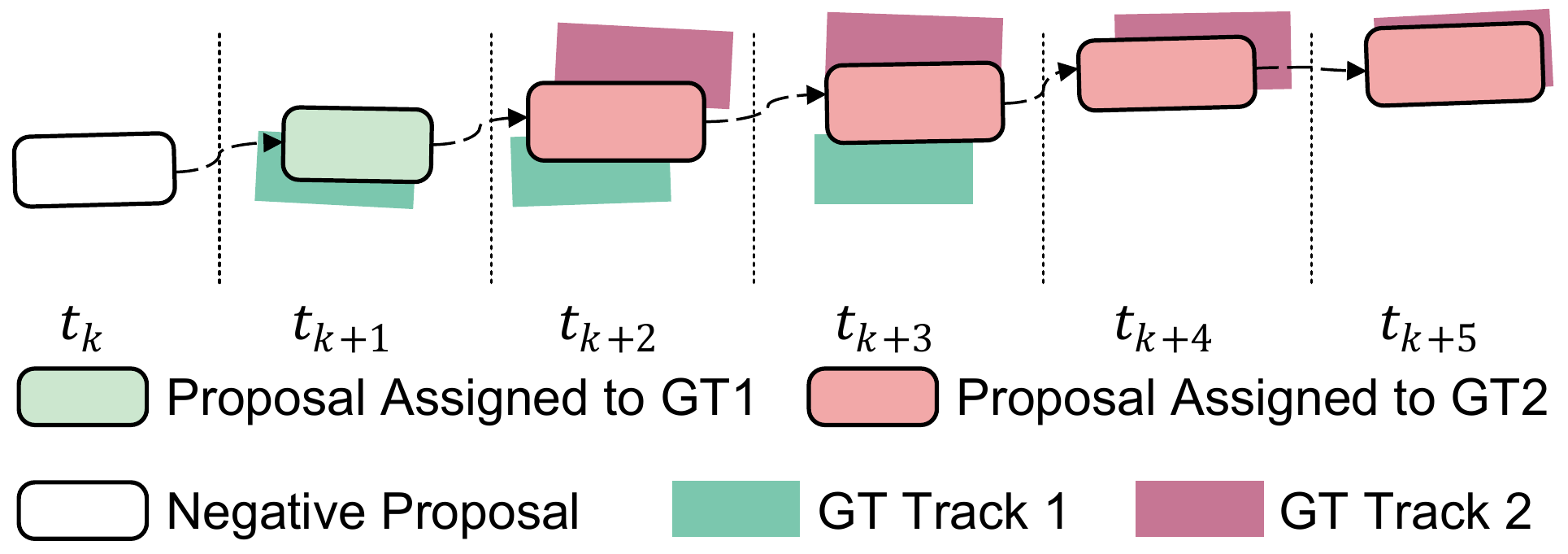}
    \caption{\textbf{The second-round assignment}.
    At $t_{k+2}$, the proposal has higher IoU with GT1, but it is assigned to GT2 due to the predicted track having higher TIoU with GT track 2.}
    \vspace{-3mm}
    \label{fig:round2}
\end{figure}

\vspace{-3mm}
\paragraph{Track IoU}
We first define Track IoU (TIoU) to measure distances between tracks, formulated in Equation~\ref{eq:track_distance}.
\begin{align}
    \text{TIoU}(T_a^{S_a}, T_b^{S_b}) = \frac{\sum_{i=1}^{|S_a \cap S_b|}{\text{IoU}(B_a^i, B_b^i)}}{|S_a \cup S_b|},
    \label{eq:track_distance}
\end{align}
where $S_a$ and $S_b$ is the indices of time step of track $T_a$ and $T_b$, respectively, and $B^i$ is the $i$-th box in the overlapped part of the two tracks.
As can be seen from Equation~\ref{eq:track_distance}, TIoU is actually the average IoU of each box pair divided by the length of the union of two tracks.
Using the TIoU, we assign the tracks in a two-round manner as follows.
\vspace{-3mm}
\paragraph{Assignment}
In the first round, we match the predicted tracks with ground-truth tracks.
A ground-truth track is assigned to a predicted track only if their TIoU is larger than a predefined threshold.
Afterward, a predicted track might have multiple ground-truth track candidates.
Predicted tracks with no candidates are regarded as negative.
In this round, the track-level matching offers better robustness and accuracy than the object-level matching.
In the second round, for a matched track pair, we assign the GT boxes to the proposals frame by frame.
Figure~\ref{fig:round2} illustrates the process.
At a certain time step, if there are multiple GT boxes from different candidate GT tracks, we simply choose one GT box from the GT track with a higher TIoU.
This second round helps alleviate the potential assignment ambiguity of the wrong associated predicted track.
\subsection{Detection Head and Losses}
\paragraph{Classification}
In the classification, we use a quality-aware loss as in~\cite{parta2, pvrcnn, fsd, gfl}.
We first calculate the IoU between each proposal and its corresponding GT.
Eventually, using the IoU, we obtain a soft classification target $q$, formulated as $q = \min(1, \max(0, 2\text{IoU} - 0.5))$.
For each proposal, we aggregate its contained point features extracted by PointNet into a proposal embedding by max pooling.
Then an MLP is adopted to predict the soft label $q$ from the embedding.
The final classification loss $\mathcal{L}_{cls}$ is the cross entropy between predicted logits and the soft labels.
\vspace{-3mm}
\paragraph{Regression}
The regression target for a positive sample is the box residual to its corresponding ground-truth box, and is the same as~\cite{parta2, fsd}.
Then the regression loss is denoted as $\mathcal{L}_{reg} = L1(\Delta_{res}, 
\widehat{\Delta_{res}})$, where $\Delta_{res}$ is the ground-truth residual and $\widehat{\Delta_{res}}$ is the predicted residual.
Similar to the classification branch, the box residual is also predicted from an aggregated proposal embedding.
\par
The total loss is defined as:
\begin{equation}
    \mathcal{L} = \mathcal{L}_{cls} + \alpha\mathcal{L}_{reg},
\end{equation}
where $\alpha$ is set to 2. More details can be found in our supplementary materials.

\subsection{Post-processing}
\label{sec:post}
\paragraph{Remove empty predictions.}
In a typical labeling process, empty bounding boxes will not be annotated even if there indeed exists an object.
Our method is capable to predict such boxes using temporal context.
To match the GT annotation process, we also remove all the empty predictions after the track-centric learning module.
\vspace{-3mm}
\paragraph{Track TTA}
In our framework, test-time augmentations (TTA)~\cite{tta} are very simple to implement.
The output tracks in different augmentations are naturally aligned according to the time steps.
Thus we conduct the simple weighted average on the predictions at each time step by scores.
For headings, we adopt the medians instead of the weighted average in case of the ``heading flip''. 
\vspace{-3mm}
\paragraph{Track Coherence Optimization (TCO)}
We can further improve the temporal coherence of tracks by object shape registration.
In particular, we specify a box containing a clear object shape, and then align other object shapes to the specified one.
We adopt the multi-way registration~\cite{multiway} via pose graph optimization for the alignment.
The pose graph has two key elements: nodes and edges. 
A node is the point cloud $P_i$ associated with a pose matrix $M_i$ which transforms $P_i$ into the specified frame point cloud.
For each edge, the transformation matrix $T_{i,j}$ that aligns the source points in $P_{i}$ to the target points in $P_{j}$. 
We use Point-to-point ICP~\cite{p2picp} to estimate the all mentioned transformations. Then the alignment is achieved by optimizing $\mathcal{M} = \{M_i\}$ via:
\begin{equation}
    \min_{\mathcal{M}} \sum_{i,j}\sum_{(p,q)\in{K}_{ij}} \|{M}_{i}p - {M}_{j}q\|_2^2.
    \label{eq:f2}
\end{equation}
In Equation~\ref{eq:f2}, ${K}_{ij}$ is the set of matched point pairs between $T_{i,j}P_i$ and $P_j$. Afterwards, we utilize the inverse of the optimized transformations to transform the box centers and heading angles in the canonical box coordinate system.
In this way, we align the poses of boxes in a track to the specified frame.
More details are in our appendices.

\section{Experiments}
\subsection{Setup}
\begin{table*}[ht]
\small
\centering

\resizebox{0.99\linewidth}{!}{
\begin{tabular}{l|c|c|c|c|c|c|c|c}
  \specialrule{1pt}{0pt}{1pt}
\toprule
\multirow{2}{*}{Methods} & \multirow{2}{*}{\shortstack[1]{\#.\\ frames}} & \multirow{2}{*}{\shortstack[1]{mAP/mAPH\\ L2}} & \multicolumn{2}{c|}{\textit{Vehicle} 3D AP/APH} & \multicolumn{2}{c|}{\textit{Pedestrian} 3D AP/APH} & \multicolumn{2}{c}{\textit{Cyclist} 3D AP/APH}\\
 & &&   L1      &   L2     &   L1      &   L2  &   L1     &   L2\\
\midrule

PV-RCNN++(center)~\cite{pvrcnnpp}&1  & 71.7/69.5 &     {79.3}/   {78.8}   &  {70.6}/{70.2}&  81.3/76.3  &  73.2/68.0 & 73.7/72.7 & 71.2/70.2\\
ConQueR~\cite{conquer}&1 & 74.0/71.6  & 78.4/77.9  & 71.0/70.5 & 82.4/76.6 & 75.8/70.1  & 77.5/76.4 & 75.2/74.1 \\
Graph-RCNN~\cite{graphrcnn}&1 & 73.2/70.9 & 80.8/80.3 & 72.6/72.1 & 82.4/76.6 & 74.4/69.0 & 75.3/74.2 & 72.5/71.5 \\
VoxelNeXt~\cite{voxelnext} & 1 & 72.2/70.1 & 78.2/77.7 & 69.9/69.4 & 81.5/76.3 & 73.5/68.6 & 76.1/74.9 & 73.3/72.2 \\
FSD~\cite{fsd}&1   &    {72.9}/   {70.8} & 79.2/{78.8} & 70.5/{70.1} &    {82.6}/   {77.3} &    {73.9}/   {69.1} &    {77.1}/{76.0} &    {74.4}/{73.3}  \\
GD-MAE~\cite{gdmae}& 1 & 74.1/71.6 & 80.2/79.8 & 72.4/72.0 & 83.1/76.7 & 75.5/69.4 & 77.2/76.2 & 74.4/73.4  \\
FlatFormer~\cite{flatformer}&3 & 73.5/72.0 & 79.7/79.2 & 71.4/71.0 & 82.0/76.1 & 74.5/71.3 & 77.2/76.1 & 74.7/73.7 \\
CenterFormer~\cite{centerformer} & 8 & 75.1/73.7 & 78.8/78.3 & 74.3/73.8 & 82.1/79.3 & 77.8/75.0 & 75.2/74.4 & 73.2/72.3\\
INT ~\cite{int} & 10 & -/73.6 & -/- & -/73.3 & -/- & -/71.9 & -/- & -/75.6 \\
MPPNet~\cite{mppnet} & 16 & 75.6/74.9 &     {82.7}/{82.3} &     {75.4}/{75.0} & 84.7/82.3 & 77.4/75.1 & 77.3/76.7 & 75.1/74.5\\
FSD++~\cite{fsd++}  & 7 &     {76.8}/{75.5} & 81.4/80.9 & 73.3/72.9 &     {85.1}/{82.2} &     {78.2}/{75.4} &     {81.2}/{80.3} &     {78.9}/    {78.1} \\
DSVT~\cite{dsvt}  & 4 & 77.5/76.2  & 82.1/81.6 & 74.5/74.1 & 86.0/83.2 & 79.1/76.4 & 81.1/80.3 & 78.8/78.0   \\
\midrule
\name w/o. any TTA & all &    83.6/82.1 & 88.0/87.3  & 81.7/81.0 &    88.9/86.1  &    83.2/80.4  &   87.7/86.7   &   85.8/84.8  \\
\name w/. Track TTA & all &     83.9/82.3 & 88.5/87.5 & 82.3/81.3 &   89.1/86.3   &   83.3/80.5   &   87.9/86.9   &  86.0/85.0   \\
\bottomrule
  \specialrule{1pt}{1pt}{0pt}
\end{tabular}
}
\caption{
    Comparison with the state-of-the-art detectors in Waymo Open Dataset validation split with the standard single-model setting.
    }
 \label{tab:wod_validation}

\end{table*}
\paragraph{Dataset}
Following pioneering work 3DAL~\cite{offboard},  we use the large-scale challenging Waymo Open Dataset (WOD)~\cite{wod} as the testbed.
WOD is currently the largest and most trustworthy dataset in the field of LiDAR-based 3D object detection, which contains 1150 sequences, 798 for training, 202 for validation, and 150 for testing.
Each sequence contains around 200 consecutive frames.
WOD adopts the strict 3D IoU as the matching metric in evaluation, which makes it a very suitable benchmark for our method since one of our objectives is to estimate accurate object poses.
\vspace{-3mm}
\paragraph{Implementation Details}
For the base detector, the training and inference schemes are consistent with the original FSD~\cite{fsd}, except that we use more input frames.
For the tracker, we adopt the default settings in Immortal Tracker~\cite{immortaltrack}.
In our track-centric learning module, the sparse UNet backbone has the same hyper-parameters as the one in FSD.
As for data augmentation, we adopt the commonly used global random flip, rotation, and scaling. We also add size and center jittering to the proposals for better robustness.
All models are trained with eight RTX 3090 GPUs with batch-size 128, which means there are 16 full-length tracks on each GPU.
Due to the limited space, we present more details in appendices.

\subsection{Main Results}
\paragraph{Comparison with state-of-the-art detectors.}

We first compare the \name with the state-of-the-art online detectors on the validation set, and the results are shown in Table~\ref{tab:wod_validation}, where \name significantly outperforms all existing methods.
For the results in the test split, the \emph{Vehicle/Pedestrian} L1 AP exceeds \textbf{90\%} with strict IoU-based evaluation metric. We present the detailed results in appendices.
\paragraph{Tracking Results}
Table~\ref{tab:wod_tracking} briefly showcases our tracking performance and also outperforms previous arts.
Detailed performance on validation and test split can be found in the appendices.
\begin{table}[H]
	\small
	\begin{center}
		\resizebox{0.99\columnwidth}{!}{
			\begin{tabular}{l|cc|cc}
				\toprule
				& 			
				\multicolumn{2}{c|}{\emph{Vehicle} } & \multicolumn{2}{c}{\emph{Pedestrian} } \\
				& MOTA$\uparrow$ & MOTP$\downarrow$  & MOTA$\uparrow$  & MOTP$\downarrow$ \\
				\midrule
				AB3DMOT*~\cite{3dmotbaseline} & 55.7 & 16.8 &  52.2 & 31.0  \\ 
                    CenterPoint*~\cite{centerpoint} & 55.1 & 16.9 & 54.9 & 31.4 \\ 
                    SimpleTrack*~\cite{simpletrack} & 56.1 & 16.8 &  57.8 & 31.3 \\
                    CenterPoint++~\cite{centerpoint}  & 56.1 & - & 57.4 & -  \\
                    Immortal Tracker~\cite{immortaltrack} & 56.4 & - &  58.2 & -   \\ \midrule
                    \name (Ours) & 71.7 & 15.0 & 70.5 & 29.3 \\
				 \bottomrule
			\end{tabular}
		}	
	\end{center}
 \vspace{-3mm}
	\caption{Tracking results on WOD validation split (L2). *: from~\cite{simpletrack}.}
    \label{tab:wod_tracking}
\end{table}
\paragraph{Comparison with the previous offline method.}
3DAL~\cite{offboard} is the most representative work in offline LiDAR-based 3D object detection.
Table~\ref{tab:comp_3dal} shows the comparison results.
\name consistently outperforms 3DAL in all metrics by large margins.
\name achieves even better results in strict IoU threshold, which indicates \name can better refine the object poses in fine grain.
\begin{table}[h]
\small
\begin{center}
\resizebox{0.99\columnwidth}{!}{
\begin{tabular}{lccccc}
  \toprule 
   \multirow{2}{4em}{IoU Threshold} & \multirow{2}{4em}{Method} &  \multicolumn{2}{c}{Vehicle} & \multicolumn{2}{c}{Pedestrian}  \\
   & & 3D & BEV & 3D & BEV \\
    \midrule
    \multirow{3}{3em}{Normal} 
  & 3DAL~\cite{offboard} & 84.5 & 93.3 & 82.9 & 86.3  \\ 
  & Ours & 88.5 & 95.5 & 89.1 & 92.3 \\
    & Improvement & +4.0 & +2.2 & +6.2 & +6.0 \\
  \midrule
    \multirow{3}{3em}{Strict}
 
  & 3DAL~\cite{offboard} & 57.8 & 84.9 & 63.7 & 75.6 \\  
  & Ours & 64.9 & 87.9 & 73.3 & 82.4  \\ 
    & Improvement & +7.1 & +3.0 & +9.6 & +6.8  \\ 
\bottomrule
 \end{tabular}
 }
\end{center}
\vspace{-5mm}
\small
\caption{\textbf{Comparison with state-of-the-art offline detectors on WOD validation split}. The normal/strict thresholds for vehicles are 0.7/0.8, and are 0.5/0.6 for pedestrians.}
\vspace{-3mm}
\label{tab:comp_3dal}
\end{table}
\subsection{Performance Inspection}
\begin{figure}[h]
    \centering
    \includegraphics[width=\linewidth]{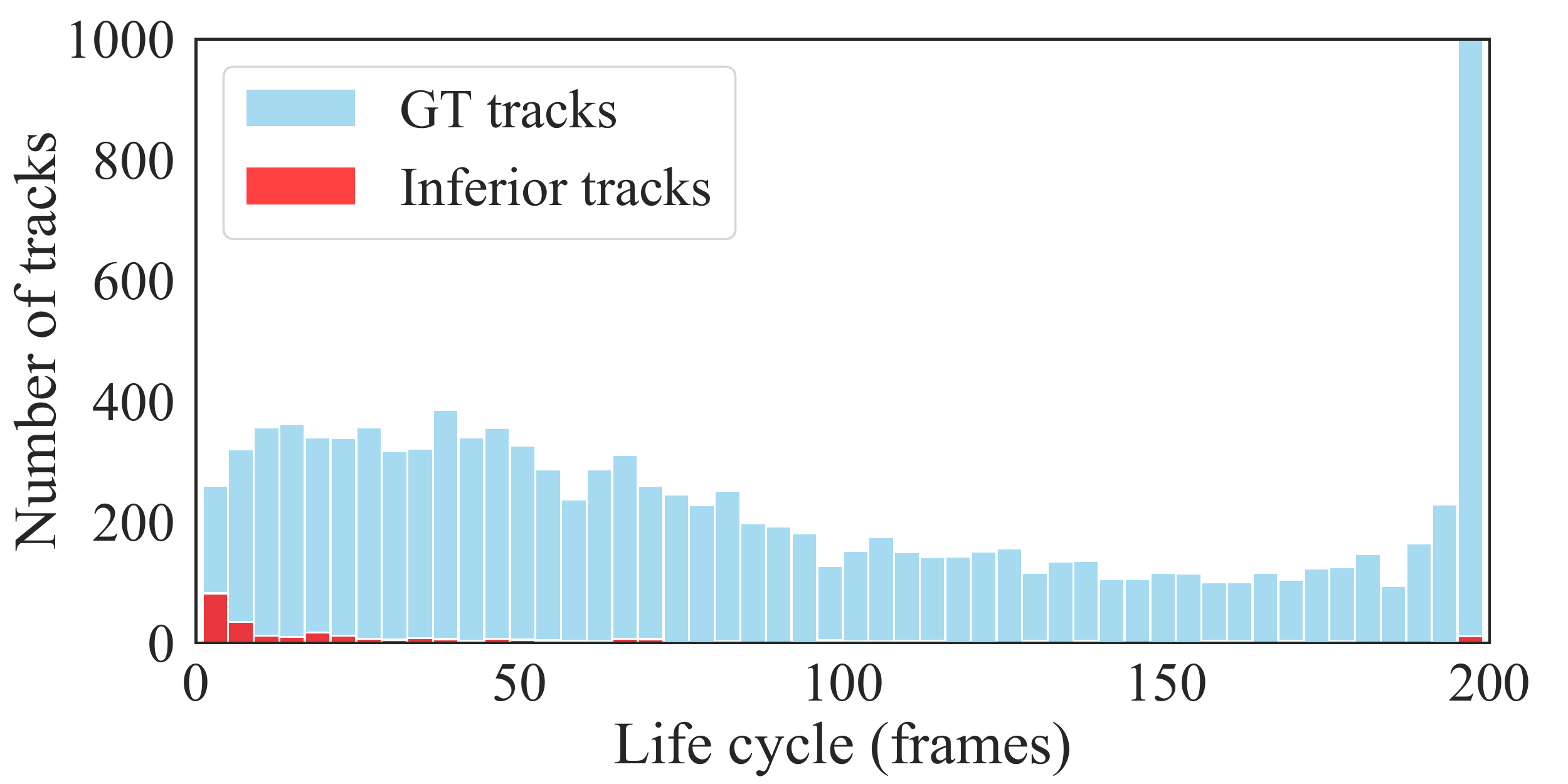}
    \caption{The life cycle histogram of normal GT tracks and inferior tracks. 
        The frame frequency is 10Hz.
    }
    \label{fig:short_tracks}
\end{figure}

\begin{table*}[h]
	\small
	\begin{center}
		\resizebox{0.99\linewidth}{!}{
                \setlength{\tabcolsep}{3pt}
			\begin{tabular}{l|cccc|cccc}
				\toprule
				& 			
				\multicolumn{4}{c|}{\emph{Vehicle} ($\sim$1.0M GTs) } & \multicolumn{4}{c}{\emph{Pedestrian} ($\sim$0.45M GTs)} \\
				&  L2 AP/APH  & T-FN$\downarrow$ & H-FP$\downarrow$ & H-TP$\uparrow$ &   L2 AP/APH  & T-FN$\downarrow$ & H-FP$\downarrow$ & H-TP$\uparrow$  \\
				\midrule
				Offline FSD & 74.9/74.4 & 21.4k (2.05\%) & 13.1k (1.25\%) & 488k (46.4\%) & 79.5/77.0 & 10.1k(2.21\%) & 6.9k (1.52\%) & 272k (59.2\%)\\
                    Refine w/. \dag  & 80.7/80.1 & 21.1k (2.00\%) & 9.2k (0.87\%) & 591k (56.3\%) & 81.0/78.3 & 10.0k (2.20\%) & 5.6k (1.21\%) & 280k (61.0\%)\\
                    Refine w/. forward ext \ddag & 81.4/80.7  & 9.1k (0.87\%) &  8.1k (0.77\%) & 594k (56.5\%) & 82.8/80.1 & 3.7k (0.82\%) & 5.4k (1.18\%) & 287k (62.7\%)\\
                    Refine w/. bidirectional ext \P & 81.7/81.0 & 5.1k (0.48\%) & 8.1k (0.77\%) & 595k (56.6\%) & 83.2/80.4 & 2.4k (0.53\%) & 5.4k (1.18\%) & 289k (63.1\%)\\
				 \bottomrule
			\end{tabular}
		}	
	\end{center}
 \vspace{-4mm}
	\caption{\textbf{Performance inspection of \namenospace}.
        TTA is not adopted.
        \dag: the proposal refinement in track-centric learning module with default tracking results from Immortal Tracker.
        \ddag: Refinement with tracks extended in the forward direction.
        \P: Refinement with tracks extended bidirectionally.
        T-FN: totally missed false negative.
        H-FP: high-confidence false positive.
        H-TP: high-precision true positive.
        The IoU threshold of H-TP is \textbf{0.9} BEV IoU for vehicles and \textbf{0.7} BEV IoU for pedestrians.
        }
	\label{tab:inspection}
\end{table*}
In this section, we thoroughly investigate our method for a comprehensive understanding in detail. 
\vspace{-3mm}
\paragraph{Overall Analysis}
We first propose several targeted metrics in addition to Average Precision to better characterize the performance.
The three metrics we proposed are the ratio of \emph{Totally missed False Negatives} (T-FNs), the ratio of \emph{High-confidence False Positives} (H-FPs), and the ratio of \emph{High-precision True Positives} (H-TPs).
Here, a T-FN is defined as the ground-truth box which does not overlap with any predictions.
To avoid being dominated by too many low-confidence predictions, we set the maximum number of predictions in each frame as 200.
H-FPs refers to false positives with confidence scores higher than a certain threshold $s_t$.
We use the score $s_t$ at 50\% recall to calibrate the different confidence distributions between models.
H-TP stands for the predicted box having high IoUs (e.g., 0.9) with the ground truth boxes.
Table~\ref{tab:inspection} presents the statistics.
Three conclusions can be drawn from it.
\begin{itemize}[leftmargin=*]
    \item The bidirectional tracking significantly reduces the T-FNs in all classes.
    In 1.03M vehicle objects, only \textbf{5.1k (0.48\%)} of them are missed by \namenospace.
    \item Although bidirectional tracking increases the number of boxes, it does not introduce extra high-confidence FP (H-FP).
    We owe it to the track learning module which gives proper confidence to proposals and effectively identifies negative samples.
    \item The refinement in the track-centric learning greatly boosts the number of high-precision predictions (H-TPs), especially for vehicles which get improved from \textbf{46.4\%} to \textbf{56.6\%}.
\end{itemize}
\vspace{-3mm}
\paragraph{Failure Cases}
We then take a closer look at those objects still missed by \namenospace.
We find the major source of missed objects are those objects with very few points.
Statistically, 50\% totally missed GTs contain less than \textbf{5 points} and 90\% of them contain less than \textbf{24 points}.
Another important characteristic of missed objects is that they are likely to belong to \textbf{short-life tracks}.
We extract GT tracks containing more than 10\% missed objects for analysis, and we refer to these tracks as \textbf{inferior tracks}. 
Fig.~\ref{fig:short_tracks} shows the life cycle distribution of all GT tracks and the inferior tracks.
As can be seen, the major part of inferior tracks have very short life cycles, which are usually shorter than \textbf{2.5 seconds}.
\begin{table*}[!t]
\vspace{-.2em}
\centering
\subfloat[
\textbf{Track length}. The maximum track length in WOD is 200. 
\label{tab:track_length}
]{
\centering
\begin{minipage}{0.2\linewidth}{\begin{center}
\tablestyle{4pt}{1.05}
\begin{tabular}{c|cc}
Lengths & Veh. & Ped. \\
 \hline
$[0, 50)$ & 77.1 & 77.6 \\
$[0, 100)$ &  78.6 & 78.0 \\
$\mathbf{[0, 200)}$ & 80.1 & 78.3 \\
\end{tabular}
\end{center}}\end{minipage}
}
\hspace{2em}
\subfloat[
\textbf{Track-centric assignment vs. object-centric assignment}.
\label{tab:track_assignment}
]{
\begin{minipage}{0.22\linewidth}{\begin{center}
\tablestyle{4pt}{1.05}
\begin{tabular}{c|cc}
Strategy & Veh. & Ped. \\
 \hline
 Object-centric & 79.6 & 78.1 \\
\textbf{Track-centric} & 80.1 & 78.3 \\
\multicolumn{3}{c}{~}\\
\end{tabular}
\end{center}}\end{minipage}
}
\subfloat[
\textbf{Separating motion states}.
\label{tab:motion}
]{
\begin{minipage}{0.22\linewidth}{\begin{center}
\tablestyle{4pt}{1.05}
\begin{tabular}{c|cc}
Strategy & Veh. & Ped.  \\
 \hline
Separated & 79.8 & 78.3 \\
\textbf{Mixed} &  80.1 &  78.3  \\
\multicolumn{3}{c}{~}\\
\end{tabular}
\end{center}}\end{minipage}
}
\centering
\hspace{5mm}
\subfloat[
\textbf{MIMO vs. MISO}. $4 \times$ means $ 4 \times$ longer training schedule.
\label{tab:mimo}
]{
\begin{minipage}{0.22\linewidth}{\begin{center}
\tablestyle{6pt}{1.05}
\begin{tabular}{c|ccc}
Mode & Veh. & Ped. & FPS\\
\hline
MISO  &  79.5 &  77.7 & 1.1\\
MISO 4$\times$  & 80.0 & 78.3 & 1.1\\
\textbf{MIMO} & 80.1 & 78.3 & 33.6\\
\end{tabular}
\end{center}}\end{minipage}
}
\hspace{2em}

\caption{
\textbf{Ablations of track-centric learning}.
We report level 2 APH for vehicle and pedestrian.
For a clean ablation, here we do not adopt track TTA and bidirectional tracking in the baseline setting.
The baseline experiments are marked in \textbf{bold}.
}
\label{tab:ablations} \vspace{-.5em}
\end{table*}
\subsection{Human Labeling Study}
\label{sec:human}
Since the annotation of accurate 3D boxes is much harder than that in 2D, human annotations usually contain errors and are inconsistent with each other. In this section, we try to compare our method with human annotations from several aspects.
\vspace{-3mm}
\paragraph{Normal Cases}
Previous art 3DAL~\cite{offboard} reports the human labeling accuracy on five sequences in the WOD validation split.
We evaluate our method on the same data, and the results are shown in Table~\ref{tab:human_3dal}.
Although 3DAL shows strong performance, it is still worse than human performance in the main metrics.
In contrast, \name consistently outperforms human performance in all the metrics by large margins.
In the strict IoU thresh (0.8), \name significantly surpasses human by \textbf{5.79 3D AP}.
\begin{table}[h]
	\small
	\begin{center}
		\resizebox{0.9\columnwidth}{!}{
			\begin{tabular}{l|cc|cc}
				\toprule
				& 			
				\multicolumn{2}{c|}{\emph{Vehicle} 3D AP } & \multicolumn{2}{c}{\emph{Vehicle}  BEV AP} \\
				& IoU=0.7  & IoU=0.8 & IoU=0.7 & IoU=0.8 \\
				\midrule
				Human & 86.45 & 60.49 & 93.86 & 86.27  \\
                    3DAL & 85.37 & 56.93 & 92.80 & 87.55\\
                    Ours & \textbf{90.19} & \textbf{66.28} & \textbf{94.26} & \textbf{89.66} \\
				 \bottomrule
			\end{tabular}
		}	
	\end{center}
 \vspace{-3mm}
	\caption{\textbf{Comparison with human performance}. Evaluation is conducted on relabeled data provided by 3DAL~\cite{offboard}.}
  \vspace{-3mm}
	\label{tab:human_3dal}
\end{table}
\vspace{-3mm}
\paragraph{Hard Cases}
\begin{figure}[h]
    \centering
    \includegraphics[width=\linewidth]{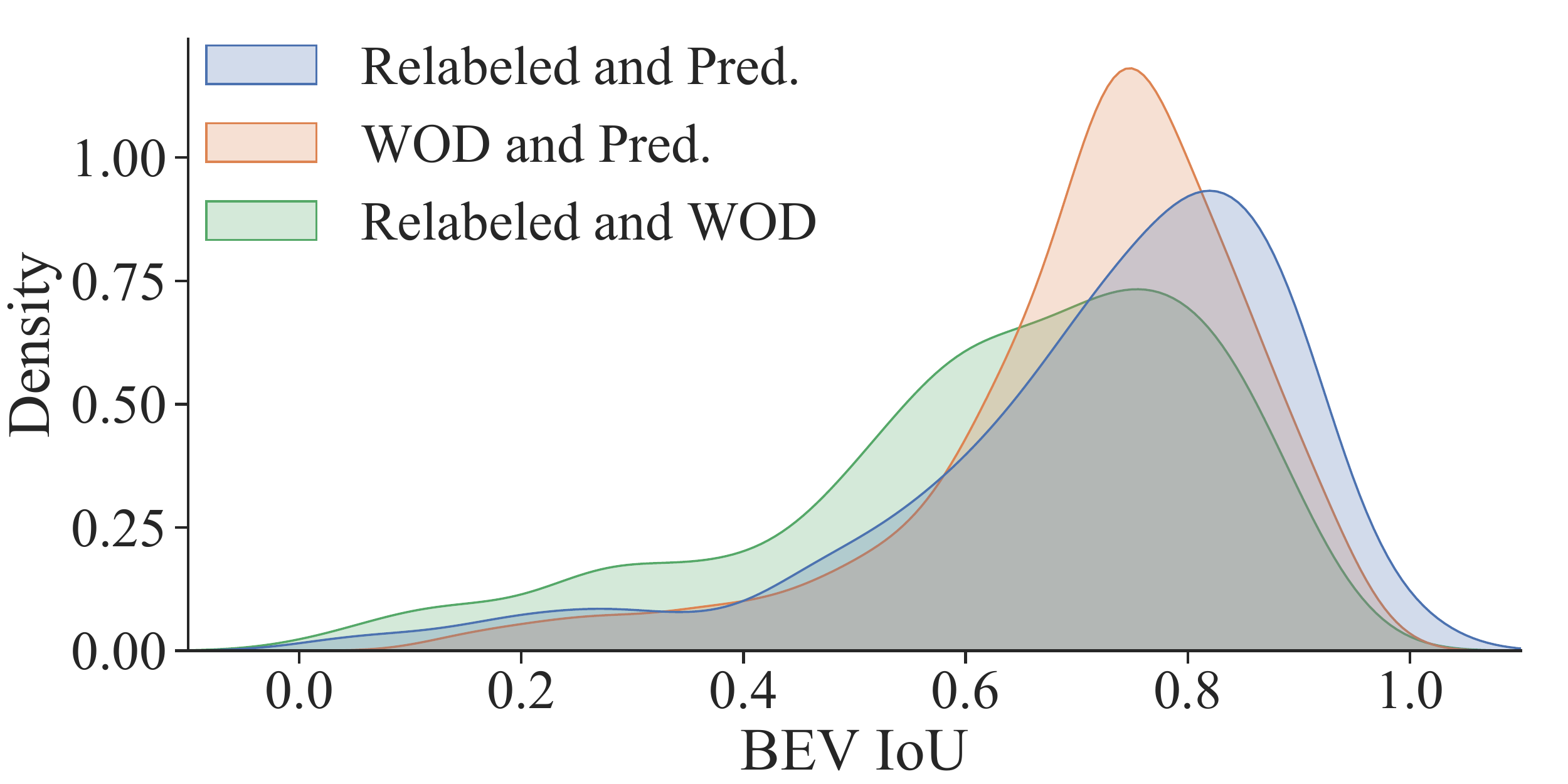}
     \vspace{-3mm}
    \caption{
    \textbf{IoU distribution between each pair of the three kinds of boxes}.
    ``Relabeled'' stands for boxes relabeled by us.
    ``WOD'' stands for official WOD annotations.
    ``Pred.'' stands for the predictions of \namenospace.
    Taking ``WOD and Pred.'' as an example, it means the distribution of IoUs between WOD official annotations and our predictions.
    The curves are smoothed for clear visualization.
    }
     \vspace{-3mm}
    \label{fig:iou_distribution}
\end{figure}
We further compare our method with human performance on the hard cases.
To this end, we ask experienced annotators to label 1000 ``hard vehicles'' in the WOD validation split carefully.
These relabeled GTs are randomly sampled from GTs which has 3D IoU with our predictions lower than 0.7 (official evaluation threshold)\footnote{We present the labeling protocol and examples of hard cases in appendices.}.
Since the box confidence given by annotators may not be accurate and have a considerable impact on the quantitative evaluation, we only calculate the IoU distribution to evaluate the results.
Note that we opt for BEV IoU instead of 3D IoU because we observed that official WOD annotations have the incorrect vertical position occasionally.
\par
After relabeling, we have three kinds of boxes: official WOD GTs, relabeled boxes, and our predictions.
Then we calculate the IoU distribution between each pair of them, shown in Figure~\ref{fig:iou_distribution}, where we draw the following two conclusions:
\begin{itemize}[leftmargin=*]
    \item \textbf{Even human annotations are not consistent for hard cases.}
    The relabeled boxes and WOD annotations have the smallest overall IoU as the green curve indicates, which means these so-called GT annotations may contain noise, especially for hard cases. 
    \item \textbf{\name has higher quality.}
    Comparing the blue curve with the green curve, our predictions have better IoU than WOD annotations on these carefully relabeled data, which indicates our method might actually have better performance than the WOD official annotations.
\end{itemize}

\subsection{Ablation Studies}
\paragraph{Input Track Length}
In our default setting, the input track length is not limited.
Here we set an upper bound for input track length to see how it affects the performance.
In Table~\ref{tab:track_length}, $[0, N]$ means that all tracks longer than $N$ will be broken into several parts.
For example, if $N$ is 50, then a 140-frame track will be broken into three tracks with lengths of 50/50/40, respectively.
The results in Table~\ref{tab:track_length} suggest that the input length is crucial and the unbounded input length offers \name a significant boost.

\paragraph{Track-centric Assignment}
We compare the proposed track-centric assignment and conventional object-centric assignment in Table \ref{tab:track_assignment}.
For the object-centric assignment, we follow the commonly-used setting and hyper-parameters in previous work~\cite{parta2, pvrcnn, fsd}, where 3D IoU is adopted as the box matching metric.
Specifically, the positive IoU threshold is 0.45 for the vehicle class and 0.35 for the pedestrian class.
The results in Table \ref{tab:track_assignment} show that track-centric assignment is more suitable in our pipeline since the track-level matching offers more robustness.

\paragraph{Motion Agnostic}
In our default setting, we do not separate static and dynamic tracks.
Here we follow the previous 3DAL~\cite{offboard} to deal with the two kinds of tracks separately.
For vehicle class, we regard a track with an average velocity higher than 1.0 m/s as dynamic.
For pedestrian class, tracks faster than 0.2 m/s are viewed as dynamic.
We train them separately and merge the results together.
Table~\ref{tab:motion} shows that separating them leads to no gains.

\paragraph{MIMO vs. MISO}
We then proceed to compare the proposed MIMO with conventional MISO.
For adapting MIMO to MISO, we choose a single proposal to supervise in every forward pass.
Table \ref{tab:mimo} shows that MISO needs a much longer training schedule to achieve similar performance to MIMO and is much slower during inference. 

\paragraph{Track Coherence Optimization (TCO)}
When given one frame high-quality human annotation, we could utilize our proposed TCO to align other predictions in the same track to this box.
Table~\ref{tab:semi} shows the results. We demonstrate consistent improvements on all metrics and all quasi-rigid classes. The improvement is especially significant when evaluated at stricter IoU.
We also present a visualization of object shapes after TCO in Figure~\ref{fig:icp_vis} in appendices to qualitatively demonstrate the effectiveness.
\begin{table}[H]
	\small
	\begin{center}
		\resizebox{0.95\columnwidth}{!}{
			\begin{tabular}{l|cc|cc}
				\toprule
				& 			
				\multicolumn{2}{c|}{\emph{Vehicle} L1 3D AP } & \multicolumn{2}{c}{\emph{Cyclist} L1 3D AP} \\
				& IoU=0.7  & IoU=0.8 & IoU=0.5 & IoU=0.6 \\
				\midrule
				\name  & 88.5 & 64.9 & 87.7 & 72.7  \\
                    +TCO &  90.2(+\textbf{1.7}) & 76.5(+\textbf{11.6}) & 88.9(+\textbf{1.2}) & 82.7(+\textbf{10.0}) \\
				 \bottomrule
			\end{tabular}
		}	
	\end{center}
 \vspace{-3mm}
	\caption{The gain from track coherence optimization on WOD validation split. One ground-truth box is given to a track for shape registration.}
	\label{tab:semi}
\end{table}

\subsection{Performance Roadmap}
As a summary, Table~\ref{tab:roadmap} shows how the performance is improved step by step from a state-of-the-art online detector to our best offline setting.
\begin{table}[H]
	\begin{center}
		\resizebox{0.99\columnwidth}{!}{
			\begin{tabular}{l|cccc}
				\toprule
			
				\multirow{2}{*}{Techniques} &\multicolumn{4}{c}{L2 3D AP/APH}  \\
				& Mean & Vehicle & Pedestrian & Cyclist \\
				\midrule
				Single-frame FSD~\cite{fsd} & 72.9/70.8 & 70.5/70.1 & 73.9/69.1 & 74.4/73.3 \\
                    Multi-frame FSD~\cite{fsd++} & 76.8/75.5 & 73.3/72.9 & 78.2/75.4 & 78.9/78.1 \\
				Offline FSD$^\ast$  & 78.8/77.5 & 74.9/74.4 & 79.5/77.0 & 82.0/81.1\\
				+Refine\dag & 81.7/80.3 & 80.7/80.1 & 81.0/78.3 & 83.3/82.5 \\
    			+Refine w/. Bi\ddag & 83.6/82.1 & 81.7/81.0& 83.2/80.4 & 85.8/84.8\\
    			+Track TTA & 83.9/82.3 & 82.3/81.3 & 83.3/80.5 & 86.0/85.0 \\
				 \bottomrule
			\end{tabular}
		}	
	\end{center}
 \vspace{-3mm}
		\caption{
            \textbf{Performance roadmap from the state-of-the-art online detector to our best offline detector.}
            $\ast$: our base detector.
            \dag: refinement with default non-extended tracking results.
            \ddag: refinement with bidirectionally extended tracking results.
  }
	\label{tab:roadmap}
\end{table}
\subsection{Runtime Evaluation}
\label{sec:runtime}
Although \name is for offline use, resource efficiency is also crucial in production.
Thus we perform a simple run-time evaluation on the previous offline method and ours, shown in Table~\ref{tab:runtime}.
\name is around 20$\times$ faster than 3DAL.
This is mainly because \name could achieve strong performance without TTA or ensemble in the base detector.
\begin{table}[H]
	\begin{center}
		\resizebox{0.85\columnwidth}{!}{
			\begin{tabular}{l|cccc}
				\toprule
				& Detection & Tracking & Refine & Total \\
				\midrule
				3DAL & 900s & 3s& 25s & 928s \\
                    \name & 36s & 4s & 6s & 46s \\
				 \bottomrule
			\end{tabular}
		}	
	\end{center}
		\caption{Time cost of processing a 20s sequence.}
	\label{tab:runtime}
\end{table}

\vspace{-2mm}
\section{Conclusion}
This paper proposed \namenospace, a high-performance offline 3D object detection system.
\name is motivated by human labeling behavior that usually first labels the easy samples in a track, and then propagates the labels to other harder samples.
Equipped with bidirectional tracking and track-centric learning, \name achieves strong detection performance, surpassing all existing models and human labeling accuracy.
Notably, only 0.48\% objects are entirely lost by \namenospace.
We adopt simple design choices in \namenospace, making it a good baseline for offline LiDAR-based auto-labeling. We will try to incorporate multi-modality input to further improve the performance in future work.


{\small
\bibliographystyle{ieee_fullname}
\bibliography{egbib}
}
\clearpage
\section*{\LARGE Appendices}
\appendix
\section{Overview}
We first present the outline of this supplementary material.
\S\ref{sec:network} contains the detailed network and input designs.
The human labeling protocol is presented in \S\ref{sec:labeling}.
The details of bidirectional tracking are in \S\ref{sec:tracking}.
\S\ref{sec:performance} offers detailed detection performance on the WOD test split and detailed tracking performance.
\S\ref{sec:scheme} elaborates our training and inference scheme.
\S\ref{sec:annos} and \S\ref{sec:hardcases} showcase some typical examples of incorrect official annotations and the sampled hard cases for relabeling.
We present the detailed algorithm of Track Coherence Optimization in \S\ref{sec:tco}.

\section{Detailed Network Structure}
\label{sec:network}
\paragraph{Input}
After tracking, for every track, we first use its containing proposals to crop points.
Each proposal is expanded by 2 meters in three size dimensions (1 meter on each side).
The expanded proposals are used to crop raw points in the corresponding frame.
In case of the out-of-memory issue caused by too many points of large objects, we randomly downsample all points in a proposal to 1024 points.
Instead, the track length is not limited.
The downsampled points are transformed into the pose of the first frame of the track and concatenated together.
We just place the points in their transformed positions without any further movement (e.g., centralizing) or editing.
We treat the concatenated points in a track as a sample in a mini-batch.

\paragraph{Track Feature Extraction}
We use the Sparse UNet in FSD~\cite{fsd} as the backbone to extract the track features (i.e., point-wise features extracted from a full track). The network hyperparameter is the same as the one in FSD, which can be accessed through their official \href{https://github.com/tusen-ai/SST/blob/main/configs/fsd/fsd_waymoD1_1x.py}{website}, we also provide our detailed configuration in the attached file with MMDetection3D format.

\paragraph{Object Feature Extraction}
As for the object feature extraction, we first use the proposal to crop points \emph{from all the concatenated track points}, and further extract the features of cropped points.
In particular, we adopt PointNet implementation in FSD to extract point features, which is more efficient than the original implementation.
We use 6 consecutive PointNets to extract the point features, and eventually, the point features of each object are aggregated by a max pooling operator.
We also provide the configuration in our attached code with detailed comments.
Since the cropped points are from different time steps, we append a binary flag to the cropped points to distinguish whether a certain point comes from the current frame.
``Current frame'' here refers to the frame to which the proposal belongs.

\section{Human Labeling Protocol}
\label{sec:labeling}
In the main paper, we conduct a human labeling study. Here we present the detailed human labeling protocol.
\paragraph{Sample Selection}
We randomly select 1000 ``hard vehicles'' to relabel.
To this end, we first calculate the maximum IoU between each GT and all predictions in the same frame.
Then we randomly sample 1000 GTs with 3D IoU lower than 0.7 and
higher than 0.1 as ``hard vehicles''.
For each selected GT, we give all the point clouds throughout its life cycle to annotators.
\paragraph{Labeling}
We ask experienced annotators to do the job with professional labeling tools.
They carefully label the object poses in three perspective views.
Given the multi-frame point clouds, they could play the clip or concatenate them to utilize the temporal information.
Although we provide point cloud sequences to annotators, they only need to label the frame containing the aforementioned ``hard vehicles''. 
Thus 1000 GTs could sufficiently cover hard cases with high diversity.

\section{Details of Bidirectional Tracking}
\label{sec:tracking}
\paragraph{Forward Tracking}
We adopt the official \href{https://github.com/ImmortalTracker/ImmortalTracker}{codebase} of ImmortalTracker to implement the forward tracking.
We make some small modifications to its default \href{https://github.com/ImmortalTracker/ImmortalTracker/blob/main/configs/waymo_configs/immortal.yaml}{setting}.
(1) We do not use the default score threshold to remove low-confidence predictions because we find it harmful to detection performance.
(2) We do not use the additional NMS since our base detector has adopted a strict NMS (IoU = 0.2).
Other key settings remain unchanged.
For example, following ImmortalTracker, we adopt 3D IoU to measure the object matching cost and a bipartite matching algorithm for object association.
During the forward tracking, we will fill in the missing predictions by a motion model which is maintained by a Kalman Filter.
All tracks longer than 100 frames are extended into the end of the sequence by the motion model.
Other tracks are extended 20 frames longer into the future.

\paragraph{Backtracing}
After forward tracking, we backtrace each track from its last frame to its first frame to estimate the motion states.
Then we backward extend the track into the past using the motion states.
Similar to the forward process, all tracks longer than 100 frames are extended into the beginning of the sequence by the motion model.
Other shorter tracks are extended 20 frames longer into the past.

\section{Detailed Performance}
\label{sec:performance}
\paragraph{Tracking Results}
Table~\ref{tab:tracking_val_detailed} and Table~\ref{tab:tracking_test_detailed} show the tracking performance of \name in Waymo Open Dataset validation and test split, respectively. 
\begin{table*}[h]
	\small
	\begin{center}
		\resizebox{0.95\linewidth}{!}{
			\begin{tabular}{l|ccc|ccc|ccc}
				\toprule
				& 			
				\multicolumn{3}{c|}{\emph{Vehicle} } & \multicolumn{3}{c|}{\emph{Pedestrian} } & \multicolumn{3}{c}{\emph{Cyclist} } \\
				& MOTA$\uparrow$ & MOTP$\downarrow$  & IDS(\%)$\downarrow$  &  MOTA$\uparrow$ & MOTP$\downarrow$  & IDS(\%)$\downarrow$  & MOTA$\uparrow$ & MOTP$\downarrow$  & IDS(\%)$\downarrow$   \\
				\midrule
				AB3DMOT*~\cite{3dmotbaseline} & 55.7 & 16.8 & 0.40 & 52.2 & 31.0 & 2.74 & -- & -- & --   \\ 
                    CenterPoint*~\cite{centerpoint} & 55.1 & 16.9 & 0.26 & 54.9 & 31.4 & 1.13 & -- & --& --\\ 
                    SimpleTrack*~\cite{simpletrack} & 56.1 & 16.8 & 0.08 & 57.8 & 31.3 & 0.42 & -- & --& --\\
                    CenterPoint++~\cite{centerpoint}  & 56.1 & - & 0.25 &  57.4 & -- & 0.94 & --& --& -- \\
                    Immortal Tracker~\cite{immortaltrack} & 56.4 & - & 0.01 & 58.2 & -- & 0.26 & -- & --& -- \\ \midrule
                    \name (Ours) & 71.2 & 15.1 & 0.0078 & 70.3 & 29.3 & 0.073 & 72.5 & 24.8 & 0.11\\
				 \bottomrule
			\end{tabular}
		}	
	\end{center}
 \vspace{-3mm}
	\caption{Tracking results on WOD validation split (L2). *:from~\cite{simpletrack}.}
    \label{tab:tracking_val_detailed}
\end{table*}
\begin{table*}[h]
	\small
	\begin{center}
		\resizebox{0.95\linewidth}{!}{
			\begin{tabular}{l|ccc|ccc|ccc}
				\toprule
				& 			
				\multicolumn{3}{c|}{\emph{Vehicle} } & \multicolumn{3}{c|}{\emph{Pedestrian} } & \multicolumn{3}{c}{\emph{Cyclist} } \\
				& MOTA$\uparrow$ & MOTP$\downarrow$  & IDS(\%)$\downarrow$  &  MOTA$\uparrow$ & MOTP$\downarrow$  & IDS(\%)$\downarrow$  & MOTA$\uparrow$ & MOTP$\downarrow$  & IDS(\%)$\downarrow$   \\
				\midrule

				InceptioLidar & 65.58 & 15.70 & 0.14 & 64.52 & 29.54 & 0.20 &65.12 & 25.42 &0.52    \\ 
                    HorizonMOT3D & 64.07 & 15.77 & 0.19 & 64.15 & 30.67 & 0.50 & 62.13 & 25.45 & 0.18 \\ 
                    MFMS\_Track & 63.14 & 15.65 & 0.07 & 63.85 & 30.19 & 0.28 & 62.83 & 25.44 & 0.69 \\
                    CasTrack & 63.66 & 15.79 & 0.05 & 64.79 & 30.24 & 0.24 & 59.34 &25.30 &0.09 \\
                    ImmortalTracker~\cite{immortaltrack} & 60.55 & 16.22 & 0.01 & 60.60 & 31.20 & 0.18 & 61.61 & 27.41 & 0.10  \\

                    \midrule
                    \name (Ours) & 74.29 & 14.26 & 0.02 & 74.21 & 28.95 & 0.05 & 71.37 & 25.18 & 0.07\\
                    \bottomrule

			\end{tabular}
		}	
	\end{center}
 \vspace{-3mm}
	\caption{Comparison with state-of-the-art online tracker on WOD test leaderboard (L2).}
    \label{tab:tracking_test_detailed}
\end{table*}
\paragraph{Detection performance on test split.}
Table~\ref{tab:detection_test_detail} showcases our detailed detection performance in the WOD test split.

\section{Training and Inference Scheme}
\label{sec:scheme}
\begin{table*}[ht]
\small
\centering

\resizebox{0.99\linewidth}{!}{
\begin{tabular}{l|c|c|c|c|c|c|c}
  \specialrule{1pt}{0pt}{1pt}
\toprule
\multirow{2}{*}{Methods}  & \multirow{2}{*}{\shortstack[1]{mAPH\\ L2}} & \multicolumn{2}{c|}{\textit{Vehicle} 3D AP/APH} & \multicolumn{2}{c|}{\textit{Pedestrian} 3D AP/APH} & \multicolumn{2}{c}{\textit{Cyclist} 3D AP/APH}\\
 &  & L1      &   L2     &   L1      &   L2  &   L1     &   L2\\
\midrule
LoGoNet\_Ens v2~\cite{logonet} & 81.96 &  	88.80/88.37 &	82.75/82.32 &  89.63/86.74 & 84.96/82.10 &     84.51/83.59	& 82.36/81.46  \\

HRI\_ADLAB\_HZ & 81.32 &  	86.77/86.40 & 80.19/79.83 & 88.59/86.01&     83.84/81.27 &     85.67/84.84 &     83.69/82.87  \\

LoGoNet\_Ens & 81.02 &  	88.33/87.87 & 82.17/81.72 &  88.98/85.96&     84.27/81.28 &     83.10/82.16 &     80.98/80.06  \\

MT-Net v2~\cite{mtnet} & 80.00 &     87.54/87.12 & 81.20/80.79 & 87.62/84.89&     82.33/79.66 &     82.80/81.74 &     80.58/79.54  \\

BEVFusion-TTA~\cite{bevfusion} & 79.97 &  	87.96/87.58 & 81.29/80.92 & 87.64/85.04&     82.19/79.65 &     82.53/81.67 &     80.17/79.33  \\

LidarMultiNet-TTA~\cite{lidarmutlinet} & 79.94 &  	87.64/87.26 & 80.73/80.36 & 87.75/85.07&     82.48/79.86 &     82.77/81.84 &     80.50/79.59  \\

MPPNetEns~\cite{mppnet}  & 79.60 &  	87.77/87.37 & 81.33/80.93 & 87.92/85.15&     82.86/80.14 &    80.74/79.90 &     78.54/77.73  \\
\midrule

\name (Ours) &  82.52 &   90.08/89.17 & 84.41/83.51 & 90.17/87.13 &     85.64/82.61 &     84.06/83.23 &     82.27/81.44  \\


\bottomrule
  \specialrule{1pt}{1pt}{0pt}
\end{tabular}
}
\caption{
    Comparison with state-of-the-art online detectors on Waymo Open Dataset test split.
    \name uses TTA but does not adopt any model ensemble.
    }
 \label{tab:detection_test_detail}

\end{table*}

\paragraph{Training Data}
For the track-centric learning module, we first generate all tracks in an offline manner.
In the whole training set, we obtain 349k vehicle tracks, 293k pedestrian tracks, and 40k cyclist tracks in total.
Since the cyclist tracks are much less than the other two categories, \textbf{we repeat the cyclist tracks by $10 \times$}.

\paragraph{Data Augmentations}
We regard an input track as a scene of the traditional detectors, so we directly adopt the default data augmentations in these detectors, including global rotation/flip/scaling/translation.
In particular, an input track is randomly rotated by $[-0.78, +0.78]$, and flipped in two axes with probability 0.5, and randomly scaled by a factor in $[0.95, 1.05]$, and randomly translated in the vertical direction by at most 0.2 meters.
In addition, we add random jittering to each input proposal.
Their centers are randomly translated by $[0.2l, 0.2w, 0.1h]$, where $l,w,h$ stand for length, width, and height, respectively.
Their widths and lengths are randomly scaled by a factor in $[0.8, 1.2]$.
The heights are randomly scaled by a factor in $[0.9, 1.1]$.
And their headings are disturbed with a maximum noise of 0.2 rad.


\paragraph{Schedule and Optimization}
We train the model for 24 epochs with a one-cycle schedule.
AdamW~\cite{adamw} is adopted as the optimizer and the maximum learning rate is 1e-3.
The model is trained on 8 RTX 3090 GPUs, with 16 samples (tracks) on each of them.
The whole training takes around 20 hours.

\paragraph{Inference}
We adopt so-called batch inference for efficiency.
32 tracks are simultaneously refined in a single forward pass, which offers us high efficiency as we demonstrate in \S\ref{sec:runtime}.

\begin{figure*}[t]
    \centering
    \includegraphics[width=0.9\linewidth]{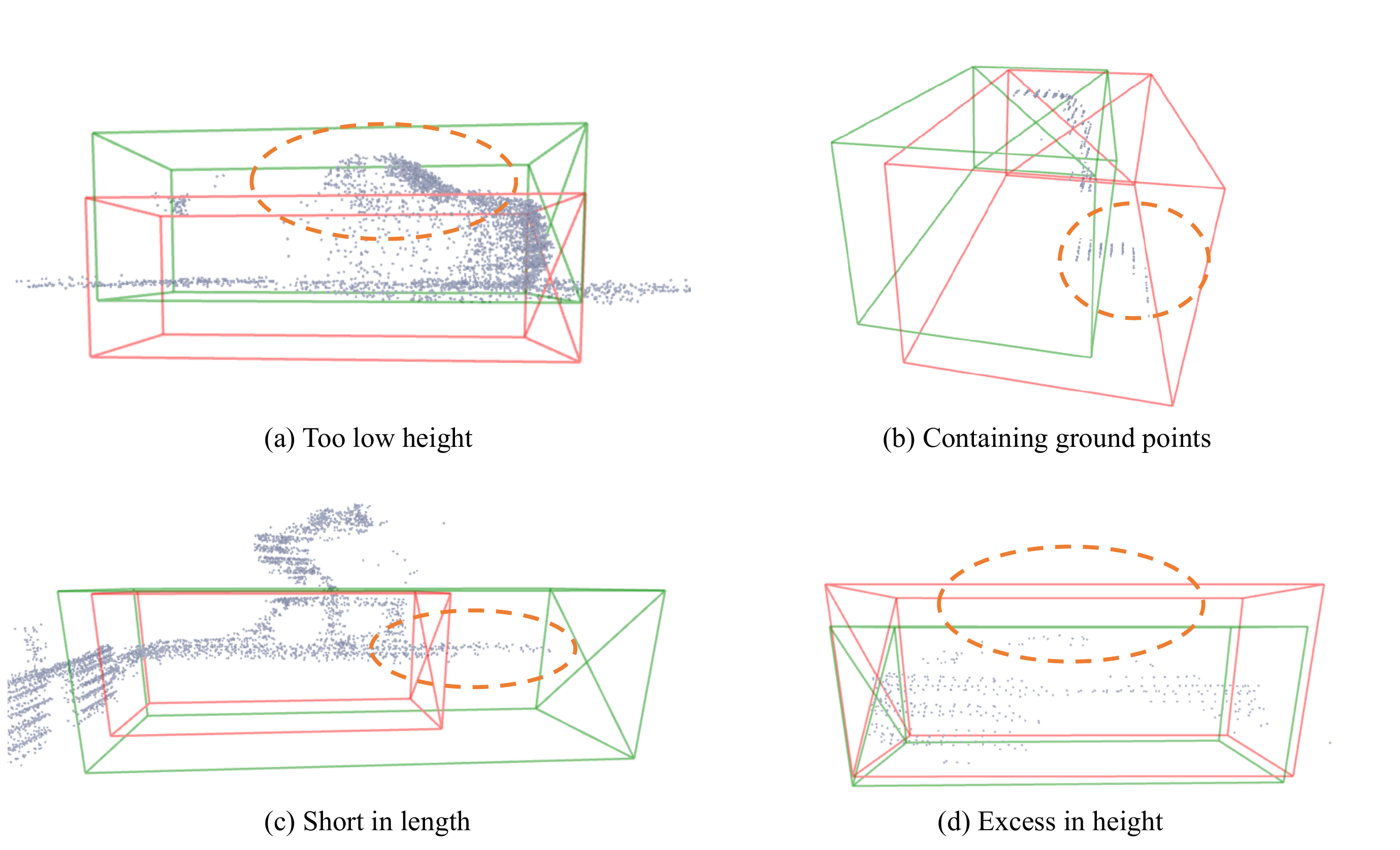}
    \caption{
    \textbf{Typical examples of incorrect official Waymo annotations.}
    The \red{red} boxes are ground-truth boxes and the \dg{green} boxes are predictions.
    We use dashed circles to indicate the incorrect parts.
    For those incorrectly annotated objects, our predictions are actually more reasonable.
    For reference, the four examples are from scenes with timestamps 1507130389559851, 1507165348439487, 1507217534286345, 1507310198254864, respectively.
    }
    \label{fig:incorrrect_annos}
\end{figure*}

\begin{figure*}[t]
    \centering
    \includegraphics[width=0.99\linewidth]{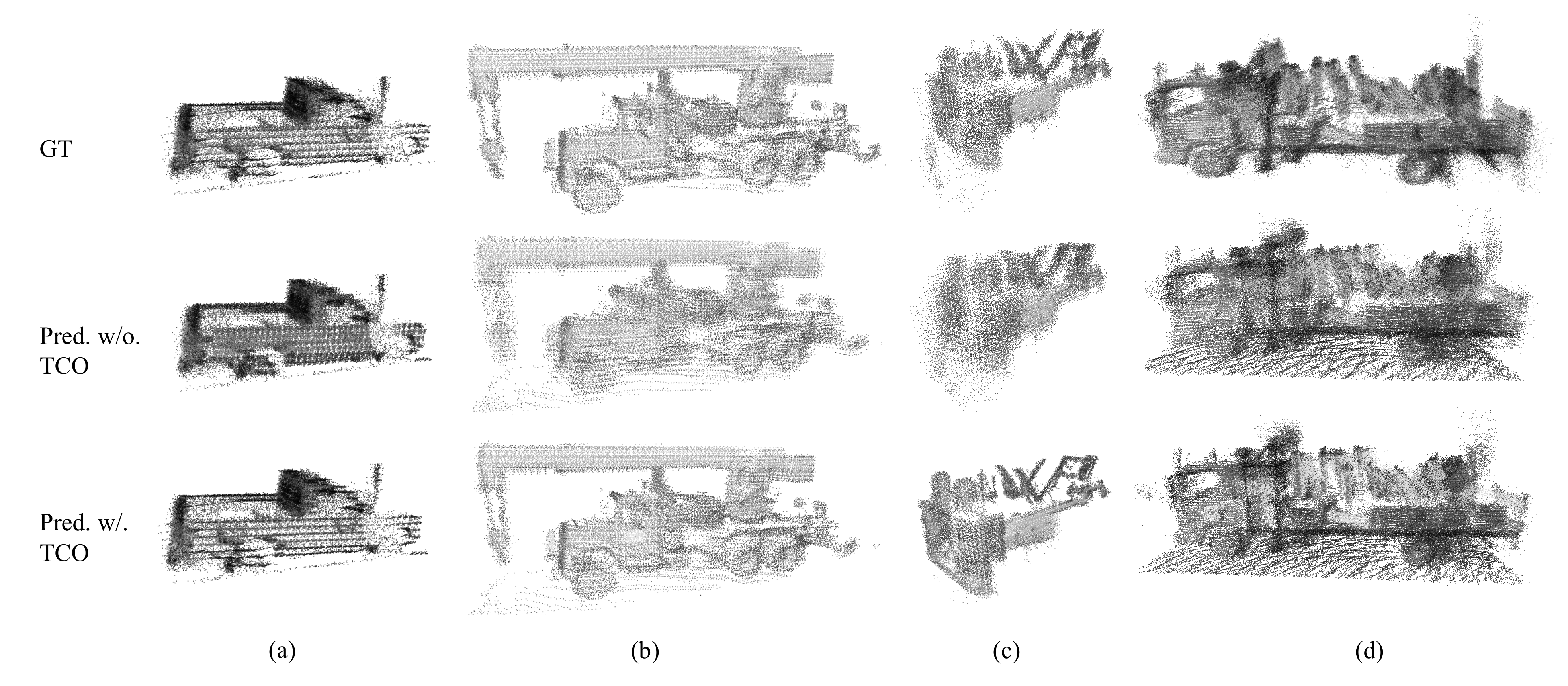}
    \caption{
    \textbf{Qualitative results of track coherence optimization (TCO).}
    We reconstruct object shapes by aligning the poses of boxes and concatenating object points from multiple frames.
    After TCO, the reconstructed shapes are much more clear, even better than the ones reconstructed by GT object poses.
    }
    \label{fig:icp_vis}
\end{figure*}

\paragraph{Track TTA}
In the main paper, we adopt conventional double-flip augmentation for the Track TTA.
Specifically, a whole input track is first horizontally (x-axis) flipped, resulting in two tracks.
The two tracks are then vertically (y-axis) flipped, resulting in four tracks in total.
In addition to the double-flip augmentation, we also try different rotations for TTA.
In particular, we adopt $[-2\pi/3, 0, +2\pi/3]$ as the rotation angles, leading to a similar performance to the double-flip strategy.
Combining double-flip augmentation and rotation further leads to around 0.1 mAP improvement.
Note that we do not adopt TTA for the base detector.



\section{Examples of Incorrect Annotations}
\label{sec:annos}
In the \S\red{5.4} of the main paper, we calculate BEV IoU instead of the 3D IoU because we find there are some incorrect annotations in the official WOD ground-truth boxes. Figure~\ref{fig:incorrrect_annos} shows some typical cases.

\section{Examples of Hard Cases}
\label{sec:hardcases}
\begin{figure*}[t]
    \centering
    \includegraphics[width=0.8\linewidth]{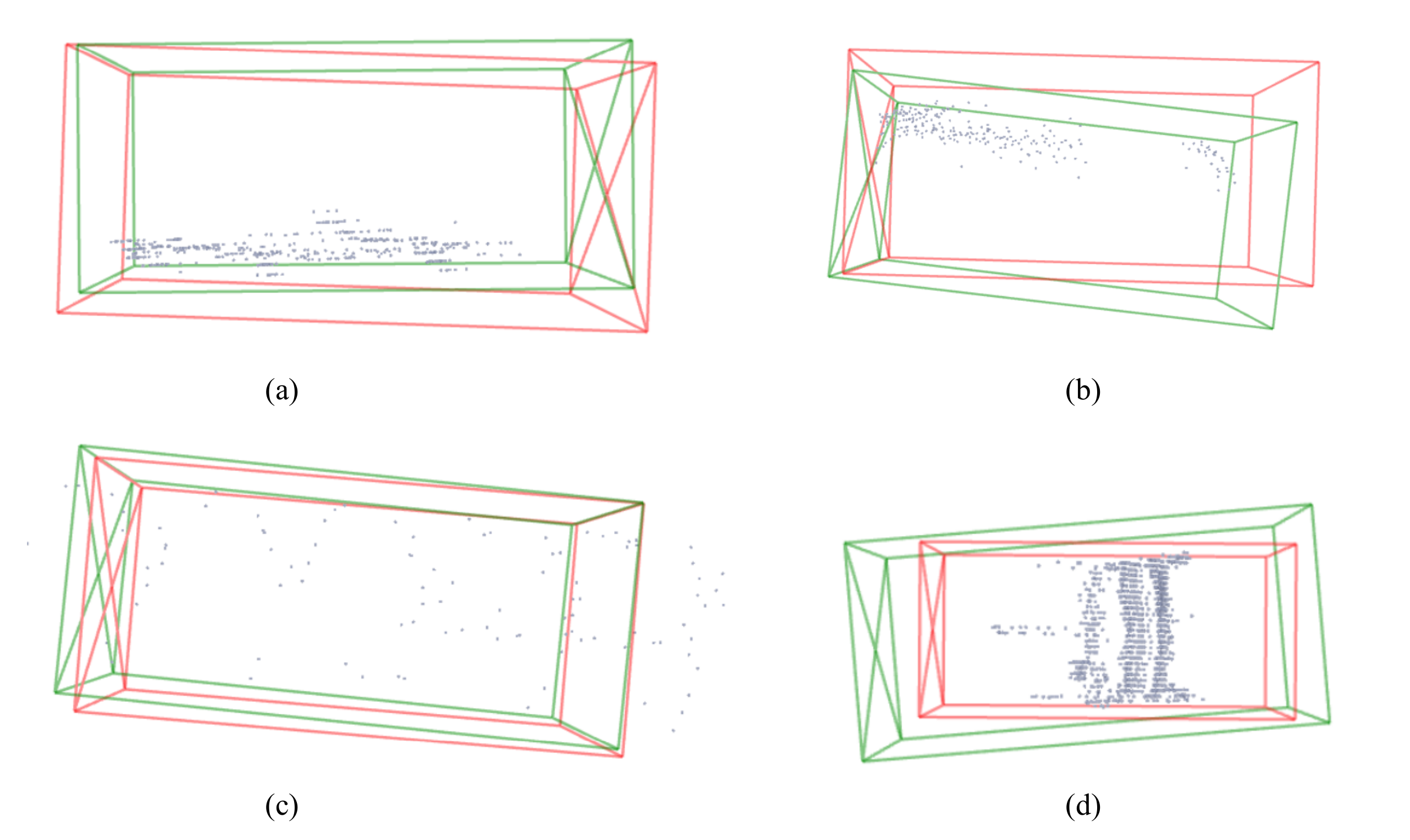}
    \caption{
    \textbf{Typical examples of our relabeled hard cases.}
    The \red{red} boxes are ground-truth boxes and \dg{green} boxes are predictions.
    They usually contain a few points or have partial shapes.
    }
    \label{fig:hard_cases}
\end{figure*}

In the \S\red{5.4} of the main paper, we randomly sample some hard cases to relabel. Figure~\ref{fig:hard_cases} qualitatively shows these cases.

\section{Details of Track Coherence Optimization (TCO)}
\label{sec:tco}
\paragraph{Overall Pipeline}
The pipeline of TCO comprises four steps: box size alignment, object shape extraction, multi-way registration, and pose quality evaluation.

\paragraph{Box Size Alignment}
The first step of TCO is to specify a frame in a track as \emph{base frame.}
For quasi-rigid objects, we assume their 3D sizes keep unchanged throughout the whole track.
So we first align all sizes in the track to the box size in the base frame.

\paragraph{Object Shape Extraction}
Afterwards, we need to extract object shapes from the complete scenes for further registration.
To this end, we first expand all proposals by 1 meter only in the height dimension, which potentially keeps more foreground points without increasing too much background clutter.
Then we use the expanded boxes to crop points in their corresponding time steps.
We use the term ``object shape'' to denote the cropped point clouds.
In a track, only frames containing more than 60 cropped points will be used for further processing. 
These cropped points (shapes) are then transformed into their canonical box coordinate.

\paragraph{Multi-way Registration}
For the next shape registration part, we adopt a multi-way registration~\cite{multiway} with pose graph optimization.
To reduce overhead, we design a sparse pose graph instead of the conventional dense pose graph.
The sparse pose graph has two key elements: nodes and sparse edges. 
A node is the point cloud (i.e., shape) $P_i$ associated with a transformation matrix $M_i$ which transforms $P_i$ into the base object shape $P_{base}$.
For each frame, we only connect it with the previous $k$ frames and the succeeding $k$ frames to construct edges, resulting in $2k$ edges.
So the constructed pose graph is sparse.
In practice, $k$ larger than 5 could achieve good performance, and we let $k = 10$ in our experiments.
For each edge, we have a transformation matrix $T_{i,j}$ aligning shape $P_{i}$ to shape $P_{j}$. 
We use point-to-point ICP~\cite{p2picp} to estimate the all mentioned transformations. 
We optimize $\mathcal{M} = \{M_i\}$ via:
\begin{equation}
    \min_{\mathcal{M}} \sum_{i,j}\sum_{(p,q)\in{K}_{ij}} \|{M}_{i}p - {M}_{j}q\|_2^2.
    \label{eq:f2}
\end{equation}
In Equation~\ref{eq:f2}, ${K}_{ij}$ is the set of point pairs between $P_i$ and $P_j$, and their pair relations are obtained by matching points in  $T_{i,j}P_i$ and points in $P_j$ with a nearest-neighbor manner.
The maximum correspondence distance of each pair is 2 meters.

\paragraph{Pose Quality Evaluation}
To avoid the failure of ICP, we employ Chamfer Distance (CD)~\cite{lidarsot} to evaluate the quality of the optimized pose for each frame.
For object shape $P_{i}$, we use CD to measure its distance to $P_{j}$:
\begin{equation}
    CD_{ij}=\frac{1}{|P_{i}|} \sum_{x \in P_{i}} \min _{y \in P_{j}}\|x-y\|_{2}+\frac{1}{|P_{j}|} \sum_{y \in P_{j}} \min _{x \in P_{i}}\|y-x\|_{2}.
    \label{eq:cd}
\end{equation}

For object shape $P_{i}$ in a track, we define its pose quality as\footnote{The Chamfer Distance here is calculated between two shapes aligned by their poses. We omit the transformation for simplicity}:
\begin{equation}
    Q_{i} = \frac{CD_{i, i-1} + CD_{i, i+1}}{2}.
\end{equation}
Then we define 
\begin{equation}
    \Delta Q_{i} = Q_{i} - Q_{i}^\prime.
\end{equation}
where $Q_{i}^\prime$ is pose quality after TCO. 
If $\Delta Q_{i}$ is positive, it means that the optimized pose becomes better.
Thus, only optimized poses with a positive $\Delta Q_{i}$ are retained, and other poses are not utilized.

Then we assume the optimal solution of Equation~\ref{eq:f2} is $\mathcal{M}^\ast = \{M_i^\ast\}$.
To get better bounding box parameters, we simply use the inverse transformation of $M_i^\ast$ to adjust the initial pose of box $B_{i}$.
We show the qualitative results of TCO in Figure~\ref{fig:icp_vis}.

\end{document}